%% file: ICRA17-SMAL-final.tex
\documentclass[letterpaper, 10 pt, conference]{ieeeconf}  % Comment this line out if you need a4paper

\IEEEoverridecommandlockouts                              % This command is only needed if
\usepackage{color}
\usepackage{times}
\usepackage{epsfig}
\usepackage{amsmath}
\usepackage{amssymb}
\usepackage{mathrsfs}
\usepackage[ruled,vlined,linesnumbered]{algorithm2e}
\usepackage{graphicx} %% This one is the package for using \includegraphics
\usepackage{subfigure}
\usepackage{subfloat}
\usepackage{url}

\title{\LARGE \bf
Sequence-based Multimodal Apprenticeship Learning For\\
Robot Perception and Decision Making
}

\author{Fei Han$^{1}$, Xue Yang$^{1}$, Yu Zhang$^{2}$, and Hao Zhang$^{1}$% <-this % stops a space
\thanks{$^{1}$Fei Han, Xue Yang and Hao Zhang are with the Department of Computer Science,
        Colorado School of Mines, 1500 Illinois Street, Golden, CO 80401, USA
        {\tt\small fhan@mines.edu, edyxueyx@gmail.com, hzhang@mines.edu}}%
\thanks{$^{2}$Yu Zhang with the Department of Computer Science and Engineering,
        Arizona State University, 699 S Mill Ave, Tempe, AZ 85281, USA
        {\tt\small yzhan442@asu.edu}}%
}

\begin{document}

\newtheorem{definition}{Definition}
\newtheorem{theorem}{Theorem}% [section]
\newtheorem{lemma}{Lemma}% [section]
\newtheorem{proposition}{Proposition}% [section]
\newtheorem{property}{Property}% [section]
\newtheorem{observation}{Observation}% [section]
\newtheorem{corollary}{Corollary}% [section]
\newtheorem{remark}{Remark}% [section]
\newtheorem{assumption}{Assumption}% [section]

\maketitle
\thispagestyle{empty}
\pagestyle{empty}

%%%%%%%%%%%%%%%%%%%%%%%%%%%%%%%%%%%%%%%%%%%%%%%%%%%%%%%%%%%%%%%%%%%%%%%%%%%%%%%%
\begin{abstract}
Apprenticeship learning has recently attracted a wide attention due to its capability of allowing robots to learn physical tasks directly from demonstrations provided by human experts.
Most previous techniques assumed that the state space is known a priori
or employed simple state representations that usually suffer from perceptual aliasing.
Different from previous research,
we propose a novel approach named Sequence-based Multimodal Apprenticeship Learning (SMAL),
which is capable to simultaneously fusing temporal information and multimodal data,
and to integrate robot perception with decision making.
To evaluate the SMAL approach, experiments are performed using both simulations and real-world robots in the challenging search and rescue scenarios.
 The empirical study has validated that our SMAL approach can effectively learn plans for robots to make decisions using sequence of multimodal observations. Experimental results have also showed that SMAL outperforms the baseline methods using individual images.

\end{abstract}

%%%%%%%%%%%%%%%%%%%%%%%%%%%%%%%%%%%%%%%%%%%%%%%%%%%%%%%%%%%%%%%%%%%%%%%%%%%%%%%%
\section{Introduction}\label{sec:introduction}

%{\color{red} \# What}

Apprenticeship learning (AL)
%, also referred to as learning from demonstration (LfD) or imitation learning (IL),
has become an active research area in robotics over the past years,
which enables a robot to learn physical tasks from expert demonstrations, without the requirement to engineer accurate task execution models.
AL has been widely applied in a variety of practical applications,
including object grasping \cite{sweeney2007model},
robotic assembly \cite{chen2003programing},
helicopter control \cite{abbeel2010autonomous},
navigation and obstacle avoidance \cite{smart2002making}, among others
\cite{argall2009survey,abbeel2004apprenticeship,bagnell2001autonomous,amit2002learning}.
% self-driving car \cite{pomerleau1991efficient}, etc.
AL methods automatically learn a mapping from world states to robot actions based on optimal or near optimal demonstrations.
These methods can also quantify the trade-off among task constraints,
which can be difficult or even impossible for manual task modeling.

%In many robotics applications, the core problem is to learn a mapping from the world state to robots' response. This mapping is also called policy. The hand-crafted policies are often very challenging because we have to quantify the trade off among various task features \cite{kolter2007hierarchical}. AL methods enables to learn the policy mechanism based on the insight that

% {\color{red} \# Challenges}\\
%Due to the great advantage of apprenticeship learning, problems in AL have been greatly studied over the last few decasde \cite{argall2009survey,abbeel2004apprenticeship,bagnell2001autonomous,amit2002learning}.
Given the advantage of AL,
however,
most previous techniques focused only on either perception or decision making without good integration between these two key components
\cite{ng2000algorithms,abbeel2004apprenticeship}.
%For example, many techniques are based on the assumption that  perception results are available for decision making \cite{ng2000algorithms,abbeel2004apprenticeship}.
It limits the capability of AL methods to address real-world problems
when a robot needs to make decisions based upon
online observations,
especially in cases when the perception data
consist of multiple modalities obtained from a variety of equipped sensors.
%while in other methods, decision making only assumed very simple features \cite{guo2003decision,he2006application}.
% % Notes from Hao: The following is not clear.
%is well designed for active perception tasks \cite{guo2003decision,he2006application}.
To address this issue, several methods were proposed to
integrate perception and planning within the same AL formulation.
A promising direction is to utilize images perceived
by robot's onboard cameras as a representation of the current state,
and then use supervised learning or reinforcement learning for decision making \cite{pomerleau1991efficient,bojarski2016end}.
However, state representation and recognition based on individual images often
suffer from the issue of perceptual aliasing (i.e., multiple distinct states of the world give rise to the same percept),
due to their incapability to incorporate temporal information or multimodal observations.
Unreliable perception will result in wrong planning and decision making, and possibly fail the tasks.

\begin{figure}[t]
\centering
\includegraphics[width=0.48\textwidth]{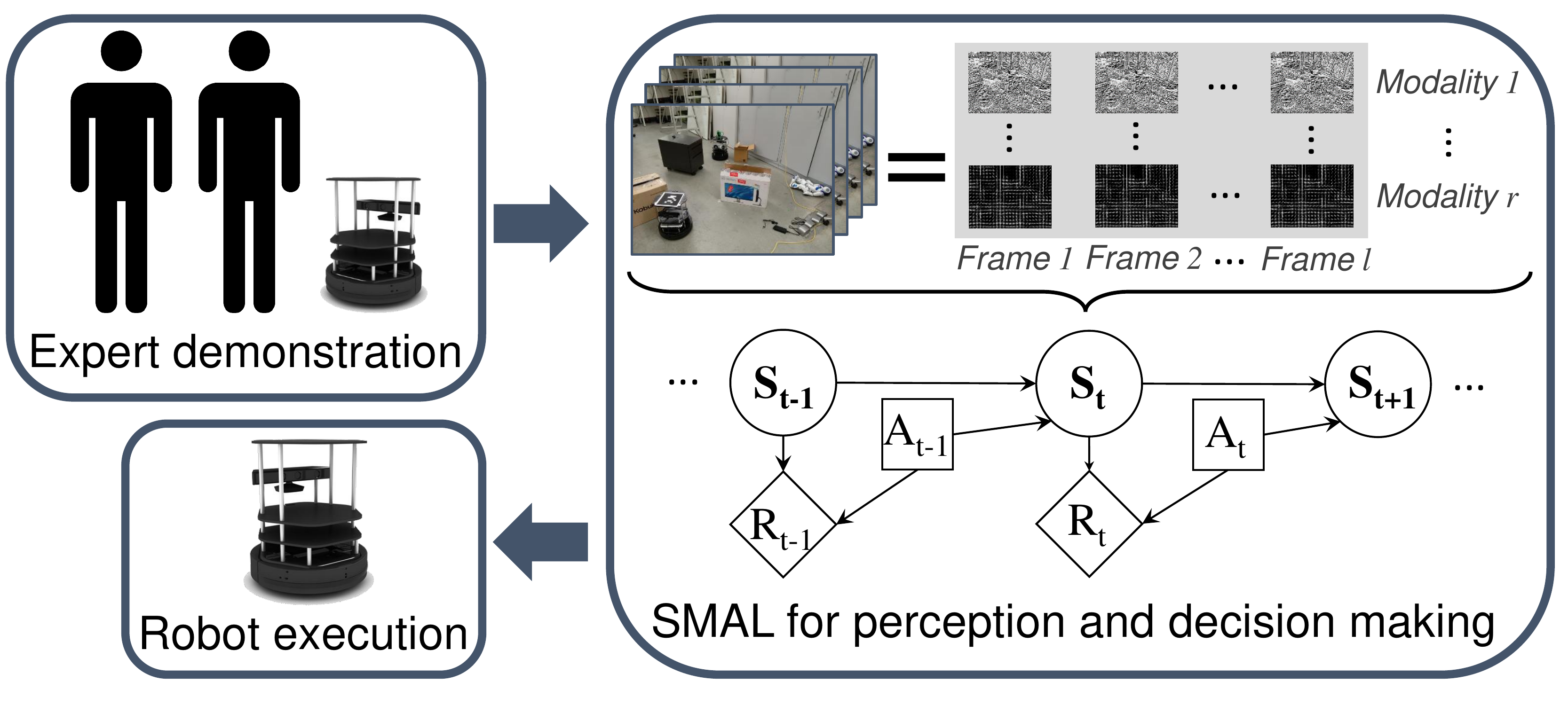}
\caption{
Overview of the proposed SMAL method to achieve robot apprenticeship learning.
Our SMAL approach is able to simultaneously integrating temporal information and multimodal observations to generate a multimodal sequence-based representation of world states.
In addition, SMAL integrates perception and decision making for robots to learn physical tasks directly from sequences of multimodal observations.
Our SMAL approach has been validated in search and rescue applications to find victims.
}\label{fig:ArchitectureCogSys}
\end{figure}

In this paper, we develop a novel \emph{Sequence-based Multimodal Apprenticeship Learning} (SMAL) method to integrate
spatio-temporal multimodal perception and decision making in the AL scenario.
Instead of using individual images,
we propose to represent a world state directly as a sequence of multimodal observations.
Then, state recognition is achieved by our new multimodal sequence-based scene matching
that integrates multimodal features obtained from each individual frame
and fuses temporal information contained in the whole sequence.
Then, we introduce a framework to integrate the sequence-based multimodal state perception with a reinforcement learning method to achieve apprenticeship learning.
We evaluate the proposed SMAL approach in challenging search and rescue applications,
as we believe our new AL paradigm has potential to address several critical tasks such as victim search and path planning.

The main contributions of this paper are twofold.
First,
we propose a novel representation of world states,
and introduce an approach to recognize the states by simultaneously fusing temporal information and multimodal data.
Second, we develop the SMAL approach that integrates multisensory robot perception and decision making to learn tasks from human experts in challenging environments with perceptual aliasing (e.g., disaster scenarios).

% Third, we investigate the performance of AL approaches in

%enable apprenticeship learning that is able to  process temporal information and multimodal data so that the perception performance can be improved.
%Second, the robot perception and decision making are able to be integrated through the SMAL method.

The rest of this paper is organized as follows.
We describe related publications in Section \ref{sec:related_work}.
In Section \ref{sec:state_learning}, we propose the sequence-based multimodal state learning.
In Section \ref{sec:SMAL}, we discuss perception and decision making integration.
After presenting experimental results in Section \ref{sec:experiment},
we conclude our paper in Section \ref{sec:conclusion}.

%\newpage
\section{Related Work}\label{sec:related_work}

In this section, we provide a review of AL techniques, and state representation and recognition methods.

\subsection{Apprenticeship Learning}

Apprenticeship learning \cite{abbeel2004apprenticeship}, also known as learning from demonstration (LfD) or imitation learning (IL) has attracted numerous attention in recent decades \cite{neu2007apprenticeship,lockerd2004tutelage,nemec2010learning},
which allows robots to accomplish tasks autonomously by learning from expert demonstrations without being told explicitly.

Many AL methods were reported in various applications,
which fall into two categories: {Direct} and {indirect approaches} \cite{neu2007apprenticeship}.
% Direct methods
\emph{Direct approaches} directly imitate experts by applying supervised learning to learn policy as a direct mapping from states to motion primitives.
% Classifier
In problems with discrete action space,
classification methods are used as mapping functions \cite{lockerd2004tutelage,dixon2004learning,chernova2007confidence,saunders2006teaching}. For example, interactive policy learning was proposed to control a car from demonstrations  based on Gaussian mixture models \cite{chernova2007confidence}.
AL techniques based on k-Nearest Neighbors (kNN) classifiers were implemented to learn obstacle avoidance and navigation
 \cite{saunders2006teaching}.
%The Bayesian method and Support Vector Machines (SVMs) were also introduced in button pressing \cite{lockerd2004tutelage} and ball sorting \cite{chernova2008teaching} tasks, respectively.
% Regression
In problems with continuous action space,
regression-based methods are typically used as state-action mapping functions
\cite{pomerleau1991efficient,vijayakumar2000locally,smart2002effective,grollman2008sparse}. For example, driving actions were learned through mapping input images to actions using neural networks  \cite{pomerleau1991efficient}.
Robot control policy was also estimated in soccer scenarios using sparse online gaussian processes \cite{grollman2008sparse}.

% Indirect methods
\emph{Indirect approaches} models the interaction between agent and environment as a Markovian decision problem,
which select the optimal policy to maximize certain reward.
% engineered rewards
Most methods manually defined the reward function.
For example, hand-crafted sparse reward functions was applied for policy synthesis in the task of corridor following in the reinforcement learning framework.
Reward functions depending on the swing angle
were implemented in a ball-in-a-cup game \cite{nemec2010learning},
in which optimal actions were chosen to maximize the accumulated reward.
% inverse reinforcement learning
Due to the great challenge to define an effective reward function \cite{argall2009survey}.
Inverse reinforcement learning was proposed to learn optimal reward functions given expert demonstrations \cite{russell1998learning,ranchod2015nonparametric,kretzschmar2016socially}. For example, three methods were demonstrated in
grid world and mountain-car tasks \cite{ng2000algorithms}.
An inverse reinforcement learning method was proposed to recover unknown reward functions under MDP framework,
which was able to output policy with performance close to that of the expert \cite{abbeel2004apprenticeship}.

However, most previous studies assume the state space is known a priori,
which still require at least partially manual construction of state space.
To address this issue,
we propose a state learning method
to automatically construct state space from multimodal sequential observations  provided in expert demonstrations.

% To address this problem, we propose a new state space learning approach through a scene recognition algorithm in search and rescue scenarios, by which the MDP model can be learned based on raw expert demonstration inputs.

%hand-crafted state space:
%\cite{saunders2006teaching}: kinematics and inner status of robots

\subsection{State Representation and Recognition}

As our objective is to integrate decision making and robot perception
that applies onboard sensors to perceive the world state,
this review will focus on methods that represent world states based on raw data directly acquired by optical cameras,
which have become a standard sensor in modern robots.

\emph{Representation:} Many techniques have been implemented to characterize and represent world states from image data based on features.
Local and global features are two main categories for visual state representation \cite{zhang2016robust}.
Local features describe local information in a part of an image,
including SIFT \cite{angeli2008fast},
ORB \cite{mur2014fast}, etc.
Such techniques apply a detector to identify interest points in an image and extract a feature vector by applying a descriptor around each interest point.
%The representations based on local features can represent discriminative scenes, and are invariant to scale, orientation and illumination changes,
%but most of them are not computational efficient.
Unlike local features, state representations based on global features describe the whole image, which encode its global color, shape, and texture signatures \cite{han2016life}.
Examples of global features
include LDB to encode intensity and gradient differences of image grid cells, GIST \cite{dymczyk2015gist} to encode dominant spatial structures,
and the recent deep feature to learn image statistics \cite{zhou2014learning}.

%Recently, deep features based on convolutional neural networks (CNNs) were also found to be able to represent scene images \cite{zhou2014learning}. These global features enables encoding the whole image information, which have shown significant performance for scene recognition \cite{milford2004ratslam,milford2012seqslam,arroyo2015towards}.

%A number of local features are reported in scene recognition, such as Scale-Invariant Feature Transform (SIFT) features \cite{angeli2008fast}, Speeded UP Robust Features (SURF) \cite{bay2006surf}, Oriented FAST and Rotated BRIEF (ORB) features \cite{mur2014fast} which based on FAST keypoint detector and BRIEF descriptor, etc.
%For example, GIST features \cite{dymczyk2015gist} represented the dominant spatial structure of a scene. The Local Difference Binary (LDB) features computed a binary string using simple intensity and gradient differences of image grid cells to represent scenes \cite{arroyo2015towards}.

\emph{Recognition:}
Most of the previous state recognition methods (e.g., scene recognition)
are based on individual-image matching,
using pairwise similarity scoring
\cite{milford2012seqslam,milford2004ratslam}, nearest neighbor search \cite{chen2006appearance,cummins2008fab,labbe2013appearance}, and sparse optimization \cite{latif2014online,yang2016enforcing}.
However, it has been demonstrated that state (or scene) recognition based on individual images cannot work well in challenging environments (e.g., with strong perceptual aliasing) \cite{arroyo2015towards,johns2013feature,milford2012seqslam,milford2004ratslam}
and fusing information from a sequence of images is critical to  match between states \cite{milford2012seqslam}.

%Single image-based scene matching approaches fail to address sequence-based scene recognition due to the fact that most similar scene templates are temporally adjacent. It has been proven that integrating information from a sequence of images is able to considerably improve the scene recognition accuracy and decrease the effect of perceptual aliasing \cite{arroyo2015towards,johns2013feature,milford2012seqslam,milford2004ratslam}.

%Most of the existing sequence-based scene matching methods compute sequence similarity by comparing all the images within the sequence with the templates and select the template sequence that has a statistically high matching score \cite{milford2012seqslam,milford2004ratslam}. However, most of the previous sequence-based methods will also have the aliasing problem without modeling the sparse nature of scene recognition.

Different from previous techniques, we propose a unified formulation
to simultaneously fuse multiple types of features to represent states and match sequence of multimodal observations for state recognition.

%our scene learning method is able to incorporate multiple feature modalities to construct a more comprehensive representation of scenes.

%In this paper, we formulate state space learning and state identification problem as a sparse sequence-based scene learning problem, which is able to model the interrelationship of the sequences and find the most representative sequence template that matches the query sequence.

%To mention a few, \cite{chen2006appearance} matched the query image with the template with the maximum similarity score using the vector distance metric. \cite{cummins2008fab} constructed a Chow Liu tree to locate the most similar scene template with respect to the query scene image.

%\cite{pomerleau1991efficient}: image input for self-driving car

\section{Sparse Multimodal State Learning}\label{sec:state_learning}
%Each MDP state $s \in \mathcal{S}$ is a unique scene observation throughout the search and rescue process. In this case, the state space $\mathcal{S}$ in the MDP model can be regarded as the scene template database (STD), which stores all unique scene observations during the data collection phase. In STD, each scene is represented by a sequence-based scene representation in our model, which has been proven to guarantee promising performance in comparison to the single image-based representation \cite{arroyo2015towards}. In this section, we formulate the sequence-based scene recognition problem as a sparse convex optimization problem.
%Besides, single image fails to incorporate temporal information, so it is not suitable to be used as state representations in our MDP model. For example, we do not know whether to approach the target or leave it if we only have a static image that captures the place near the target. However, this information is encoded if a sequence of images is used as the representation.

We propose a novel SMAL approach to (1) represent and recognize states based on multimodal observation sequences,
and (2) integrate state learning with decision making
to guide robot actions
(e.g., performing victim search and rescue in disaster areas).
This section focuses on contribution (1),
and contribution (2) will be detailed in the Section \ref{sec:SMAL}.

\emph{Notation.} In this paper, we represent vectors as boldface lowercase letters, and matrices using boldface, capital letters.
Given a matrix $\mathbf{M} = \{m_{ij}\}\in\mathbb{R}^{n\times m}$,
we refer to its $i$-th row and $j$-th column as $\mathbf{m}^{i}$ and $\mathbf{m}_{j}$, respectively.
The $\ell_1$-norm of a vector $\mathbf{v} \in \mathbb{R}^n$  is defined as $\|\mathbf{v}\|_1 = \sum_{i=1}^{n} |v_i|$,
and the $\ell_2$-norm of  $\mathbf{v}$  is defined as $\|\mathbf{v}\|_2 = \sqrt{\mathbf{v}^\top\mathbf{v}}$.
The $\ell_{2,1}$-norm of the matrix $\mathbf{M}$ is defined as:
\begin{eqnarray}\label{eq:l21norm}
\|\mathbf{M}\|_{2,1} =  \sum_{i=1}^{n} \sqrt{\sum_{j=1}^{m}{m_{ij}^{2}}} = \sum_{i=1}^{n} \|\mathbf{m}^i\|_2
\end{eqnarray}

\subsection{Sequence-based Multimodal State Matching}

To solve the problem of state identification in challenging real-world environments (e.g., disaster scenarios in search and rescue operations),
we propose to incorporate a temporal sequence of observations (e.g., images) for state recognition and fuse multiple heterogenous sensing modalities to capture comprehensive environmental information to address perceptual aliasing.

Assume a set of templates encoding the states (e.g., scenes in victim search) from a target area $\mathbf{X} = [\mathbf{x}_1, \mathbf{x}_2, \cdots, \mathbf{x}_n]\in\mathbb{R}^{m\times n}$,
and each template contains a set of $r$ heterogenous feature modalities $\mathbf{x}_i = \left[(\mathbf{x}_i^1)^\top, (\mathbf{x}_i^2)^\top, \cdots, (\mathbf{x}_i^r)^\top\right]^\top\in\mathbb{R}^{m}$,
where $\mathbf{x}_i^j\in\mathbb{R}^{m_j},j=1,\cdots,r$ represents the feature vector of length $m_j$ extracted from the $j$-the feature modality and $m = \sum_{j=1}^r m_j$.
Because our method focuses on sequence-based state learning,
we group adjacent observations (e.g., camera frames) together as a temporal sequence to encode each state,
resulting in the set of sequence-based templates $\mathbf{X} = [\mathbf{X}^1, \mathbf{X}^2, \cdots, \mathbf{X}^k]$,
where  $\mathbf{X}^j,1\leq j\leq k$ denotes the $j$-th sequence that contains $l$ images acquired in a short time period,
and $k$ is the number of sequences in the set satisfying  $k = \lfloor n/l\rfloor$.
Then, given a new query sequence containing a set of $l$
multimodal observations $\mathbf{Y} = [\mathbf{y}_1, \mathbf{y}_2, \cdots, \mathbf{y}_l]\in\mathbb{R}^{m\times l}$,
we formulate state identification as a learning task to estimate the weight matrix, $\mathbf{W} = [\mathbf{w}_1, \mathbf{w}_2, \cdots, \mathbf{w}_l]$:
%, using regularized sparse optimization:
\begin{align}
\mathbf{W} = \begin{bmatrix}
     \mathbf{w}_{1}^{1} & \mathbf{w}_{2}^{1} & \dots  & \mathbf{w}_{l}^{1} \\
     \mathbf{w}_{1}^{2} & \mathbf{w}_{2}^{2} & \dots  & \mathbf{w}_{l}^{1} \\
     \vdots & \vdots & \ddots & \vdots \\
     \mathbf{w}_{1}^{k} & \mathbf{w}_{2}^{k} & \dots  & \mathbf{w}_{l}^{k}
\end{bmatrix}
\in \mathbb{R}^{n \times l} \;,
\end{align}
where $\mathbf{w}_p^q\in\mathbb{R}^{l}$ denotes the weights of the templates in the $q$-th sequence $\mathbf{X}^q$ with respect to the $p$-th query observation $\mathbf{y}_p$
in the sequence $\mathbf{Y}$.

Since individual observations in template and query sequences can be noisy or contain missing values,
we propose to constrain each observation $\mathbf{y}$ in the sequence $\mathbf{Y}$ to only rely on a small number of representative template sequences for state recognition,
leading to the regularized sparse optimization problem as follows:
\begin{align}\label{eq:sigma_obj}
\min_\mathbf{W} \sum_{i=1}^l \left(\|\mathbf{X}\mathbf{w}_i - \mathbf{y}_i\|_2 + \lambda\|\mathbf{w}_i\|_1 \right),
\end{align}
where the $\ell_1$-norm regularization of $\mathbf{w}_i$ forces the sparsity of the scene templates used to represent the query scene.
Eq. (\ref{eq:sigma_obj}) can be rewritten as a more compact matrix expression:
\begin{align}\label{eq:matrix_obj_0}
\min_\mathbf{W} \|(\mathbf{X}\mathbf{W} - \mathbf{Y})^\top\|_{2,1} + \lambda\|\mathbf{W}\|_1,
\end{align}
where $\|\mathbf{W}\|_1 = \sum_{i=1}^l\|\mathbf{w}_i\|_1$.

However, the regularizer of weight matrix $\mathbf{W}$ in Eq. (\ref{eq:matrix_obj_0}) is an element-wise $\ell_1$-norm,
which ignores the interrelationship among individual feature modalities within each observation $\mathbf{y}$.
%Thus, the objective function  cannot encode the group structure of observation sequences,
% and is incapable to match the query sequence with template sequences.
To encode this interrelationship among individual modalities within an individual observation $\mathbf{y}$, we use the $\ell_{2,1}$-norm as a new regularization:
\begin{align}\label{eq:matrix_obj_1}
\min_\mathbf{W} \|(\mathbf{X}\mathbf{W} - \mathbf{Y})^\top\|_{2,1} + \lambda\|\mathbf{W}\|_{2,1}.
\end{align}
The $\ell_{2,1}$-norm regularization applies an $\ell_2$-norm to enforce group effects of all individual modalities in the same individual observation,
and uses an $\ell_1$-norm to enforce the sparsity among individual observations.
%That is, all individual observations in $\mathbf{Y}$ have a similar weight for the same template while the templates are sparse.

To enable sequence-based state recognition,
we propose a new regularization to model the group structure among all sequences. We name it the $S_1$-norm, because it is a structured
$\ell_1$-norm encoding the group structure of $\mathbf{W}$, as follows:
\begin{align}
\|\mathbf{W}\|_{S_1} = \sum_{i=1}^l\sum_{j=1}^k \|\mathbf{w}_i^j\|_2.
\end{align}
The $S_1$-norm applies the $\ell_2$-norm on individual observations within each sequence,
and the $\ell_1$-norm among sequences.
That is, the new $S_1$-norm not only enforces the observations within the same sequence to have similar weights, but also enforces the sparsity between sequences.
For example, if a template sequence $\textbf{X}^i$ is not representative for a query observation $\mathbf{Y}$, the weights of the individual observations in $\textbf{X}^i$ have small values; otherwise, their weights are large.

Thus, the final optimization problem becomes
\begin{align}\label{eq:obj}
\min_\mathbf{W} \|(\mathbf{X}\mathbf{W} - \mathbf{Y})^\top\|_{2,1} + \lambda_1\|\mathbf{W}\|_{2,1} + \lambda_2\|\mathbf{W}\|_{S_1}.
\end{align}

%The combination of $\ell_{2,1}$ and $S_1$-norms are able to deal with sequence misalignment, since individual observations that are similar to the current state but not in the most discriminative template sequence can be activated.

\subsection{State Space Learning and State Identification}

Our previous discussion is based upon the assumption that the state space has been provided during the training phase using expert demonstrations.
However, the critical problems of how to construct state space has not been discussed.

To address this problem in the training phase,
we introduce a new approach in Algorithm \ref{alg:state_space_learning} to automatically construct the state space $\mathcal{S}$ for our sequence-based state recognition.
% utilized in SMAL.
Intuitively,
if a query sequence does not match any template sequences within the database, it will be inserted into the database.
Formally,
after obtaining the optimal weight matrix $\mathbf{W}$,
given a new sequence $\mathbf{Y}$ during training,
we identify its state by matching $\mathbf{Y}$ with all existing template sequences $\mathbf{X}$.
If the weight of a template sequence $\mathbf{X}^j$ satisfies:
\begin{align}\label{eq:match}
\sum_{i=1}^l\|\mathbf{w}_i^j\|_1 \leq \tau,
\end{align}
where $\tau$ is a threshold with a small value, then we conclude that $\mathbf{Y}$ does not match the sequence $\mathbf{X}^j$.
If $\mathbf{Y}$ does not have any matches in $\mathbf{X}$,
we add $\mathbf{Y}$ into the template database.
This approach ensures that there exists only one representative sequence in the template database to encode the same state (e.g., the same scene with similar viewpoints).
If duplicated sequences are provided,
our algorithm will ignore them, and the state space $\mathcal{S}$ will remain the same.

\begin{algorithm}[htp]
\SetAlgoLined
\SetKwData{Left}{left}\SetKwData{This}{this}\SetKwData{Up}{up}
\SetKwFunction{Union}{Union}\SetKwFunction{FindCompress}{FindCompress}
\SetKwInOut{Input}{Input}\SetKwInOut{Output}{Output}
\SetNlSty{textrm}{}{:}
\SetKwComment{tcc}{//}{} % define comment style
% \SetKwComment{tcc}{/*}{*/} % define comment style

% \small{

\Input{
Observations recorded during demonstrations
       }
\Output{
    $\mathcal{S}$ (state space), $\mathbf{X}$ (state template database, or STD), and $s\text{-}stream$ (state stream).
    }
\BlankLine

Initialize:
$\mathbf{X}, \mathcal{S}, s\text{-}stream = \varnothing$.
%number of videos is $n_v$, number of frames in video $i$ is $f_i$
%video index $v_i = 1, v_i \in \{1, 2, \cdots, n_v \}$,
%frame index in each video $f_{ij} = 1, j \in \{1, 2, \cdots, n_{f}\}$;

%$\slash\ast$ Establish the template scene database $TSD$ and state transition map $STM$. $\ast\slash$

    \While {there exist unprocessed observations}
    {
%    		Calculate multimodal features for current $l$ frames from video $i$ as a query observation $\mathbf{Y}$ ;
%		Obtain the current sequence of observations $\mathbf{Y}$;
    		
    		Calculate the optimal weight matrix $\mathbf{W}$ according to Algorithm \ref{alg:OptAlgorithm} with respect to $\mathbf{X}$ and the current sequence of observations $\mathbf{Y}$;
    		
        \uIf {no match is found by Eq. (\ref{eq:match})}
        {
            $\mathbf{X} \leftarrow [\mathbf{X}, \mathbf{Y}]$;\\
            Add the new state to the state space $\mathcal{S}$;
        }
        \Else
        {
        		Find the matched state by Eq. (\ref{eq:scene_identify});
        }

        Append the current state to $s\text{-}stream$;

        Go to the next sequence of observations;

     }

\Return $\mathcal{S}, \mathbf{X}, s\text{-}stream$.

\caption{State space learning} \label{alg:state_space_learning}

\end{algorithm}

During the execution phase,
given the query sequence of multimodal observations $\mathbf{Y}$ obtained by the robot,
our SMAL method recognizes its state by solving the following problem:
\begin{align}\label{eq:scene_identify}
s = \arg\max_j\sum_{i=1}^l\|\mathbf{w}_i^j\|_1, j = 1, 2, \cdots, k,
\end{align}
where $\mathbf{w}_i^j$ is computed by Algorithm \ref{alg:OptAlgorithm}.

\subsection{Optimization Algorithm}\label{sec:sub:optimization}

Although the optimization problem in Eq. (\ref{eq:obj}) is convex,
it is challenging to solve it since there are non-smooth terms in the objective function. Here we provide an efficient algorithm to solve this problem that grantees theoretical convergence.

After taking the derivative of Eq. (\ref{eq:obj}) with respect to $\mathbf{W}$ and setting it to $0$, we have
\begin{align}
\mathbf{X}^\top\mathbf{X}\mathbf{W}\mathbf{P} - \mathbf{X}^\top\mathbf{Y}\mathbf{P}
+ \lambda_1\mathbf{Q}\mathbf{W} + \lambda_2\mathbf{R}^i \mathbf{W} = \mathbf{0},
\end{align}
where $\mathbf{P}$ is a diagonal matrix with the $i$-th diagonal element equals $p_{ii}=\frac{1}{2\|\mathbf{y}_i-\mathbf{X}\mathbf{w}_i\|_2}$, $\mathbf{Q}$ is a diagonal matrix with the $i$-th element as $\frac{1}{2\|\mathbf{w}^i\|_2}$, and $\mathbf{R}^i$ is a block diagonal matrix with the $i$-th diagonal block as $\frac{1}{2\|\mathbf{w}_i^j\|_2}\mathbf{I}$, where $\mathbf{I}$ denotes an $l$ dimensional identity matrix. For each $i$, we obtain
\begin{align}
p_{ii}\mathbf{X}^\top\mathbf{X}\mathbf{w}_i - p_{ii}\mathbf{X}^\top\mathbf{y}_i
+ \lambda_1\mathbf{Q}\mathbf{w}_i + \lambda_2\mathbf{R}^i \mathbf{w}_i = \mathbf{0}.
\end{align}
Therefore, $\mathbf{w}_i$ can be calculated by
\begin{align}\label{eq:w_i}
\mathbf{w}_i = p_{ii}\left(p_{ii}\mathbf{X}^\top\mathbf{X} + \lambda_1\mathbf{Q}
+ \lambda_2\mathbf{R}^i\right)^{-1} \mathbf{X}^\top\mathbf{y}_i.
\end{align}

We can observe that the matrices $\mathbf{P}, \mathbf{Q},$ and $\mathbf{R}$ in Eq. (\ref{eq:w_i}) all depend on the weight matrix $\mathbf{W}$, which are unknown.
To solve this regularized optimization problem, we propose an iterative solver as presented in Algorithm \ref{alg:OptAlgorithm}.
We can prove that Algorithm \ref{alg:OptAlgorithm} guarantees the theoretical convergence to the global optimal solution. Detailed analysis and mathematical proof is provided in Appendix.

%In addition, since the solution in this algorithm has a closed form in each iteration,
%Algorithm \ref{alg:OptAlgorithm} converges very fast, which can be demonstrated in both simulation and real-world robotic applications presented in Section \ref{sec:experiment}.

\begin{algorithm}[htp]
\SetAlgoLined
\SetKwData{Left}{left}\SetKwData{This}{this}\SetKwData{Up}{up}
\SetKwFunction{Union}{Union}\SetKwFunction{FindCompress}{FindCompress}
\SetKwInOut{Input}{Input}\SetKwInOut{Output}{Output}
\SetNlSty{textrm}{}{:}
\SetKwComment{tcc}{//}{} % define comment style
% \SetKwComment{tcc}{/*}{*/} % define comment style

% \small{

\Input{
The scene templates $\mathbf{X}\in\mathbb{R}^{m\times n}$,\\
$\,\;$the query sequence of frames $\mathbf{Y}\in\mathbb{R}^{m\times l}$.
       }
\Output{
    The weight matrix $\mathbf{W}\in \mathbb{R}^{n\times l}$.
    }
\BlankLine
% \emph{special treatment of the first line}
Initialize $\mathbf{W}\in \mathbb{R}^{n\times l}$;

\While {not converge}
{
Calculate the diagonal matrix $\mathbf{P}$ with the $i$-th diagonal element as $p_{ii} = \frac{1}{2\|\mathbf{y}_i - \mathbf{X}\mathbf{w}_i\|_2}$;

Calculate the diagonal matrix $\mathbf{Q}$ with the $i$-th diagonal element as $\frac{1}{2\|\mathbf{w}^i\|_2}$;

Calculate the block diagonal matrix $\mathbf{R}^i$ $(1\leq i \leq l)$ with the $j$-th diagonal block as
$\frac{1}{2\|\mathbf{w}_i^j\|_2}\mathbf{I}_j$;

For each $\mathbf{w}_i$ ($1 \leq i \leq s$), calculate
$\mathbf{w}_i = {p}_{ii} \left({p}_{ii} \mathbf{X}^\top \mathbf{X} + \lambda_1 \mathbf{Q} + \lambda_2 \mathbf{R}^i\right)^{-1} \mathbf{X}^\top \mathbf{y}_i$;
}

\Return $\mathbf{W}\in \mathbb{R}^{n\times l}$.

% } %\small

\caption{An iterative algorithm to solve the sparse optimization problem in Eq. (\ref{eq:obj}).}\label{alg:OptAlgorithm}

\end{algorithm}

\section{Integration of State Perception and \\Decision Making }\label{sec:SMAL}
Beyond the ability to automatically learn states,
our SMAL method is also able to integrate state perception and decision making.
This integration allows a robot to directly utilize raw multisensory observation sequences to make decisions and take actions,
without assuming
perfect perception or hand-crafted states that are not practical in complicated real-world environments (e.g., search and rescue scenarios).

%Then, the focus of this section is to integrate state perception
%with decision making to allow robot to take actions
%The decision is made according to the state the robot thought it is in so that it can finish the tasks as demonstrations by experts.

%Our framework is more significant in real world applications including search and rescue, since the autonomous search and rescue robot has to make appropriate decisions based on its observation to find the victims.

We propose to achieve the integration of our state perception with the general Markov decision process (MDP) model,
which has been widely employed for robot decision making,
to show the generalization of our SMAL method that has the potential to impact
various robotics applications using MDP.
From the viewpoint of real-world online robot execution,
the input data into our integrated model is the raw multimodal observation sequences obtained by sensors equipped on the robot,
and the output of our SMAL method is an optimal action learned in response to the state identified by our perception method.
%The output of the perception system is the current state matching through sequence-based multimodal representation.
%After the robot executes the action, it observes the surroundings and identify the new state again. This procedure is conducted repeatedly unless it reaches the end state, which indicates that the target has been found and the task ends.

Formally, the integrated perception and decision making model of our SMAL method is represented as a tuple $\Omega = (\mathcal{S}, \mathcal{A}, T, R, \gamma)$,
%(i.e., same notation as MDP), where
% S
where $\mathcal{S} = \{ s_0, s_1, \cdots, s_{N_s}\}$ denotes a finite set of discrete states;
% A
$\mathcal{A} = \{ a_0, a_1, \cdots, a_{N_a}\}$ represents a finite set of discrete actions that human/robot can perform to activate state transitions;
% T
$T: \mathcal{S} \times \mathcal{A} \times \mathcal{S} \rightarrow [0, 1]$ denotes a discrete transition function representing the probability of a state transition resulted from an action;
% R
$R: \mathcal{S} \times \mathcal{A} \rightarrow \mathbb{R}$ denotes a mapping from the state-action pair to a scalar, representing the immediate reward received when the robot takes action $a \in \mathcal{A}$ in state $s \in \mathcal{S}$;
% gamma
and $\gamma \in [0,1]$ is a reward discount factor.
Different from previous MDP-based methods whose states are typically computed at a specific time point and represented by a single modality,
our integrated model represents a state based on a sequence of observations with multimodal modalities.
This integration is realized using our sequence-based multimodal state recognition method that transfers a multimodal observation sequence $\mathbf{Y} \in\mathbb{R}^{m\times l}$ into a discrete value $s = s(\textbf{Y}) \in \mathbb{Z}$, as defined in Eq. (\ref{eq:scene_identify}).

Same as all MDP-based decision making, our integrated model aims to learn a policy $\pi$ that is defined as a mapping from the learned state space $\mathcal{S}$ to the action space $\mathcal{A}$.
The value of a policy $\pi$ is given by $V^\pi = \sum_{t=0}^\infty \gamma^t R(s_t,\pi(s_t))$.
Then, the objective of decision making is to find an optimal policy $\pi^\star$ to maximize the value function $V^\pi$:
\begin{align}
\pi^\star = \arg\max_{\pi}\sum_{t=0}^\infty \gamma^t R(s_t,\pi(s_t))
\end{align}

In the following, we describe our implemented methods to learn other components of the MDP model used in integrated SMAL method, as follows:

%Though we already learn the state space $\mathcal{S}$ by Algorithm \ref{alg:state_space_learning}, there are still other components in the MDP modeling that need to be learned.

%The robot keeps the same atom movement between two image frames captured by camera. For example, the autonomous robot keeps moving forward until the next image frame is captured by the camera.
%% human experts
%For human experts, there are kinematic sensors equipped in human's helmet (GPS, IMU, depth sensor, etc) to store human expert's translation and rotation movements. We use these kinematic information to learn the action space similar to that by teleoperated robot agent.

\emph{Learning Action Space.}
The action space $\mathcal{A}$ can be learned based on the kinematic data collected during expert demonstrations.
%There are two types of expert agents: {humans} and {teleoperated robots}.
%% teleoperated robots
In our experiments,
teleoperation command streams provided by humans are recorded and used to learn the action space $\mathcal{A}$,
where each action $a\in\mathcal{A}$ consists of a sequence of $l$ {atom movements}.
Such atom movements include {moving forward, moving backward, turning left, and turning right}.
Ideally, actions are continuous,
but robots perform actions in discrete-time during the execution phase, since the specific optimal action is selected based on the current state, which is discrete and recognized at every $l$ frames.
The action space $\mathcal{A}$ is learned by Algorithm \ref{alg:action_space_learning}.

% Each action $a\in\mathcal{A}$ with respect to state $s\in\mathcal{S}$ is composed by $l$ \emph{atom movements}.
%For example, the frame rate of the camera is assumed to be 10fps, and the sequence length $l = 10$, which means the robot identifies scenes in every 10 frames (1 second). The robot action $a$ with respect to state $l$ is a sequence of $10$ \emph{atom movements}, i.e. $a = \{jjjjjiiiii\}$, where $j$ represents rotating left and $i$ denotes moving forward.

\begin{algorithm}[htp]
\SetAlgoLined
\SetKwData{Left}{left}\SetKwData{This}{this}\SetKwData{Up}{up}
\SetKwFunction{Union}{Union}\SetKwFunction{FindCompress}{FindCompress}
\SetKwInOut{Input}{Input}\SetKwInOut{Output}{Output}
\SetNlSty{textrm}{}{:}
\SetKwComment{tcc}{//}{} % define comment style
% \SetKwComment{tcc}{/*}{*/} % define comment style

% \small{

\Input{
Recorded kinematic stream $k\text{-}stream$, and
state stream $s\text{-}stream$ learned by Algorithm \ref{alg:state_space_learning}
       }
\Output{
    The action space $\mathcal{A}$, and action stream $a\text{-}stream$.
    }
\BlankLine

Initialize:
$a\text{-}stream, \mathcal{A} = \varnothing$.
%number of videos is $n_v$, number of frames in video $i$ is $f_i$
%video index $v_i = 1, v_i \in \{1, 2, \cdots, n_v \}$,
%frame index in each video $f_{ij} = 1, j \in \{1, 2, \cdots, n_{f}\}$;

%$\slash\ast$ Establish the template scene database $TSD$ and state transition map $STM$. $\ast\slash$

    \While {there exists unprocessed kinematic data}
    {
    		Get a sequence of $l$ atom movements $am$ from the kinematic stream
    		
    		Append $am$ to the action stream $a\text{-}stream$;
    		
    		\If {$am$ is not contained in $\mathcal{A}$}
    		{
    			Insert $am$ to $\mathcal{A}$;
    		}

     }

\Return $\mathcal{A}, a\text{-}stream$.

\caption{Algorithm to learn action space $\mathcal{A}$}\label{alg:action_space_learning}

\end{algorithm}

\textit{Learning State Transition.}
The state transition $T(s,a,s')$
represents the probability that the system will end up in state $s'$ after taking action $a$ in state $s$.
The state transition $T$ is learned using the state and action streams obtained in Algorithms \ref{alg:state_space_learning} and \ref{alg:action_space_learning}, respectively.
In our implementation, the state transition is learned by Algorithm \ref{alg:state_transition_learning}.

\begin{algorithm}[htp]
\SetAlgoLined
\SetKwData{Left}{left}\SetKwData{This}{this}\SetKwData{Up}{up}
\SetKwFunction{Union}{Union}\SetKwFunction{FindCompress}{FindCompress}
\SetKwInOut{Input}{Input}\SetKwInOut{Output}{Output}
\SetNlSty{textrm}{}{:}
\SetKwComment{tcc}{//}{} % define comment style
% \SetKwComment{tcc}{/*}{*/} % define comment style

% \small{

\Input{
State stream $s\text{-}stream$ and action stream $a\text{-}stream$
       }
\Output{
    The state transitions $T$
    }
\BlankLine

Initialize:
State transition map $STM = \varnothing$,

%$\slash\ast$ Establish the template scene database $TSD$ and state transition map $STM$. $\ast\slash$

\For {$i = 1 : \text{length of }s\text{-}stream$}
{
    Append the value of key $s(i)$ with $\left(a(i), s(i+1)\right)$.
}

%$\slash\ast$ Calculate the state transition $T$. $\ast\slash$

\For {key $s$ in $STM$}
{
    $T(s,a,s') = \dfrac{\text{Number of } (a,s') \text{ in } STM[s]}{\text{Number of } (a) \text{ in } STM[s]}$
}

\Return $T$.

\caption{Algorithm to learn state transition.}\label{alg:state_transition_learning}

\end{algorithm}

\textit{Learning Immediate Reward.}
After the MDP model $\Omega = (\mathcal{S}, \mathcal{A}, T)\backslash R$ is learned, we are able to learn the immediate reward $R(s,a)$ provided by the human demonstrations and a predefined $\gamma$.
A widely used technique is inverse reinforcement learning.
We directly employed the technique in \cite{ng2000algorithms},
in which reward learning is formulated as a sparse optimization problem since the maximum reward (i.e., finding victims) in our application is achieved at the end state.

%\subsection{Inverse Reinforcement Learning}
%The objective of inverse reinforcement learning is to find immediate reward functions $R(s,a),s\in\mathcal{S},a\in\mathcal{A}$ to model demonstration effects or demonstrator goals when applied to a MDP model.
%
%A number of inverse reinforcement learning methods have been reported \cite{ng2000algorithms,abbeel2004apprenticeship,herman2015inverse,audiffren2015maximum}. In \cite{ng2000algorithms}, The reward learning is formulated as a sparse optimization problem based on the fact that only a large reward was assigned when the agent achieves the end state. This sparse reward formulation can be widely applied in search and navigation scenarios, and it is used in this paper to learn the immediate reward function as part of our SMAL system.
%
%
%
%Our learning system includes two modules: State learning module and MDP model learning module.
%The main advantage of our approach is that only videos showing expert demonstrations are needed as the system input, and we do not acquire the detailed state information as well as other information in the MDP model, including state transition, reward function, etc. Our system is able to learn the MDP model that fits the input expert demonstration well. After the system is established, we are able to apply it to autonomous robots in search and rescue scenarios so that they can perform the similar task as experts.

\input{experiment}

\section{Conclusion}\label{sec:conclusion}
We propose a novel \emph{sequence-based multimodal apprenticeship learning} approach
that can automatically learn and identify world states,
and integrates perception and decision making.
The SMAL approach represents each state as a sequence of multimodal observations by simultaneously fusing
temporal information and multimodal data.
The SMAL approach also integrates robot perception and decision making to learn tasks from human demonstrations to enable effective robot actions in challenging environments with perceptual aliasing.
To evaluate the performance of the SMAL method,
experiments using both simulations and real-world robots
are performed in the challenging search and rescue applications.
Qualitative results have validated that our method is able to
guide autonomous robots to successfully finish the search and rescue task.
In addition, quantitative evaluation results have demonstrated
that our SMAL method outperforms baseline methods based on individual images to find victims in the challenging search and rescue applications.

\appendices
\section{Convergence Analysis of Algorithm \ref{alg:OptAlgorithm}}\label{appendix}

\begin{theorem}\label{thm:opt}
Algorithm \ref{alg:OptAlgorithm} decreases the objective value of the problem in Eq. (\ref{eq:obj}) in each iteration.
\end{theorem}

The following lemma \cite{nie2010efficient} is used to prove Theorem \ref{thm:opt}.

\begin{lemma}\label{lemma}
For any nonzero vector $\mathbf{\tilde{a}}$ and $\mathbf{a}$, the following inequality holds:
% \begin{eqnarray}
$\|\mathbf{\tilde{a}}\|_2 - \frac{\|\mathbf{\tilde{a}}\|_2^2}{2\|\mathbf{a}\|_2}
\leq
\|\mathbf{a}\|_2 - \frac{\|\mathbf{a}\|_2^2}{2\|\mathbf{a}\|_2}
$.
%\end{eqnarray}
\end{lemma}

Then we are ready to prove the convergence of Algorithm \ref{alg:OptAlgorithm}, which is represented by Theorem \ref{thm:opt}.

\begin{proof}
We denote the update of $\mathbf{W}$ is $\mathbf{\tilde{W}}$.
According to Step 6 in Algorithm \ref{alg:OptAlgorithm},
we have:
{
\small
\begin{align}
\mathbf{\tilde{W}} &= \underset{\mathbf{W}} {\mathrm{arg\,min}}\;
Tr((\mathbf{XW} - \mathbf{Y})\mathbf{P}(\mathbf{XW}-\mathbf{Y})^\top) \nonumber
\\
& + \lambda_1 Tr(\mathbf{W}^\top \mathbf{QW}) + \lambda_2 \sum_{i=1}^{l} \mathbf{w}_i^\top \mathbf{R}^i \mathbf{w}_i.
\end{align}
% \vspace{-12pt}
}
Thus, we can obtain
{
\small
\begin{align}
& Tr((\mathbf{X\tilde{W}} - \mathbf{Y})\mathbf{P}(\mathbf{X\tilde{W}}-\mathbf{Y})^\top) \nonumber \\
& \qquad + \lambda_1 Tr(\mathbf{\tilde{W}}^\top \mathbf{Q\tilde{W}}) + \lambda_2 \sum_{i=1}^{l} \mathbf{\tilde{w}}_i^\top \mathbf{R}^i \mathbf{\tilde{w}}_i \nonumber \\
&\leq Tr((\mathbf{XW} - \mathbf{Y})\mathbf{P}(\mathbf{XW}-\mathbf{Y})^\top) \nonumber \\
& \qquad + \lambda_1 Tr(\mathbf{W}^\top \mathbf{QW}) + \lambda_2 \sum_{i=1}^{l} \mathbf{w}_i^\top \mathbf{R}^i \mathbf{w}_i
\end{align}
}
We are able to derive the following inequalities according to the definition of $\mathbf{P}$, $\mathbf{Q}$, and $\mathbf{R}$:
{
\small
\begin{align}
&\sum_{i=1}^{l} \left(
\frac{\|\mathbf{X}\mathbf{\tilde{w}}_i - \mathbf{y}_i\|^2_2}{2\|\mathbf{X}\mathbf{w}_i - \mathbf{y}_i\|_2}
+ \lambda_1 \frac{\|\mathbf{\tilde{w}}\|_2^2}{2\|\mathbf{w}\|_2}
+ \lambda_2 \sum_{j=1}^{k} \frac{\|\mathbf{\tilde{w}}_i^j\|^2_2}{2\|\mathbf{w}_i^j\|_2}
\right) \nonumber \\
&\leq
\sum_{i=1}^{l} \left(
\frac{\|\mathbf{X}\mathbf{{w}}_i - \mathbf{y}_i\|^2_2}{2\|\mathbf{X}\mathbf{w}_i - \mathbf{y}_i\|_2}
+ \lambda_1 \frac{\|\mathbf{{w}}\|_2^2}{2\|\mathbf{w}\|_2}
+ \lambda_2 \sum_{j=1}^{k} \frac{\|\mathbf{{w}}_i^j\|^2_2}{2\|\mathbf{w}_i^j\|_2}
\right) \nonumber
\vspace{-6pt}
\end{align}
}
According to Lemma \ref{lemma},
we obtain the inequalities:
{
\small
\begin{align}\label{eq:UVWderive}
& \sum_{i=1}^{l} \left(
\|\mathbf{X}\mathbf{\tilde{w}}_i - \mathbf{y}_i\|_2
- \frac{\|\mathbf{X}\mathbf{\tilde{w}}_i - \mathbf{y}_i\|^2_2}{2\|\mathbf{X}\mathbf{w}_i - \mathbf{y}_i\|_2}
\right) \nonumber \\
&\leq
\sum_{i=1}^{l} \left(
\|\mathbf{X}\mathbf{{w}}_i - \mathbf{y}_i\|_2
- \frac{\|\mathbf{X}\mathbf{{w}}_i - \mathbf{y}_i\|^2_2}{2\|\mathbf{X}\mathbf{w}_i - \mathbf{y}_i\|_2}
\right)
\end{align}
\vspace{-6pt}
\begin{align}
\sum_{i=1}^{l} \left(
\|\mathbf{\tilde{w}}\|_2 - \lambda_1 \frac{\|\mathbf{\tilde{w}}\|_2^2}{2\|\mathbf{w}\|_2}
\right)
\leq
\sum_{i=1}^{l} \left(
\|\mathbf{{w}}\|_2 - \lambda_1 \frac{\|\mathbf{{w}}\|_2^2}{2\|\mathbf{w}\|_2}
\right) \nonumber
\end{align}
\vspace{-8pt}
\begin{align}
\sum_{i=1}^{l}\sum_{j=1}^{k} \left( \|\mathbf{\tilde{w}}_i^j\|_2 - \frac{\|\mathbf{\tilde{w}}_i^j\|^2_2}{2\|\mathbf{w}_i^j\|_2} \right)
\leq
\sum_{i=1}^{l}\sum_{j=1}^{k} \left( \|\mathbf{{w}}_i^j\|_2 - \frac{\|\mathbf{{w}}_i^j\|^2_2}{2\|\mathbf{w}_i^j\|_2} \right) \nonumber
\end{align}
}
After computing the summation of the three equations in Eq. (\ref{eq:UVWderive}) on both sides (weighted by $\lambda$s), we obtain:
{
\small
\begin{align}\label{eq:proof}
&\sum_{i=1}^{l}
\|(\mathbf{X}\mathbf{\tilde{w}}_i - \mathbf{y}_i)^\top\|_{2} +
\lambda_1 \|\mathbf{\tilde{w}}\|_{2} +
\lambda_2 \|\mathbf{\tilde{w}}\|_{2}  \nonumber\\
&\leq
\sum_{i=1}^{l}
\|(\mathbf{X}\mathbf{{w}}_i - \mathbf{y}_i)^\top\|_{2} +
\lambda_1 \|\mathbf{{w}}\|_{2} +
\lambda_2 \|\mathbf{{w}}\|_{2}
\end{align}
}
Thus, we conclude that Algorithm \ref{alg:OptAlgorithm} decreases the objective value monotonically during each iteration.
Because Eq. (\ref{eq:obj}) is a convex optimization function,
Algorithm \ref{alg:OptAlgorithm} converges to the global optimal solution.
\end{proof}

\bibliographystyle{IEEEtran}
\bibliography{ref_abbr}

\end{document}

%% file: experiment.tex
\section{Experiments}\label{sec:experiment}
To evaluate the performance of our SMAL approach,
we performed two sets of experiments in different scenarios to address the application of robot-assisted search and rescue,
including
(1) urban search and rescue in simulation,
and (2) indoor search tasks using real robots.
The mission objective for the robot is to find a victim within the environment,
who are not directly viewable by the robot.

In our experiments, the (simulated and real) robots employ a camera to perceive the surrounding world;
multiple feature modalities are applied to extract information to represent the world.
To enable real-time performance,
we intentionally use feature modalities that can be extracted efficiently, including low-resolution color features on $24\times 32$ downsampled images and  histogram of oriented gradients features on $240\times 320$ downsampled images.
The visual feature modalities are normalized and concatenated as a multimodal representation of individual observations.

%Both qualitative and quantitative results are provided,
%and comparison with baseline methods .

%The SMAL approach was implemented using Matlab.
%%including the Robotics System Toolbox on Linux laptops.
%For the Webots simulation, we used a Linux laptop with an i5 2.7 GHz CPU, 16G memory without GPU acceleration. In the experiment using real TurtleBot, its own preinstalled laptop ({\color{red} \# to make sure i3 2.0 GHz CPU, 8G memory)} was utilized in all steps, including data collection, model learning and robot control.

\begin{figure*}[htp]
\subfigure[Simulated environments]{
    \label{fig:csm_webots_map} %% label for second subfigure
    \begin{minipage}[b]{0.22\textwidth}
      \centering
        \includegraphics[height = 1.475in]{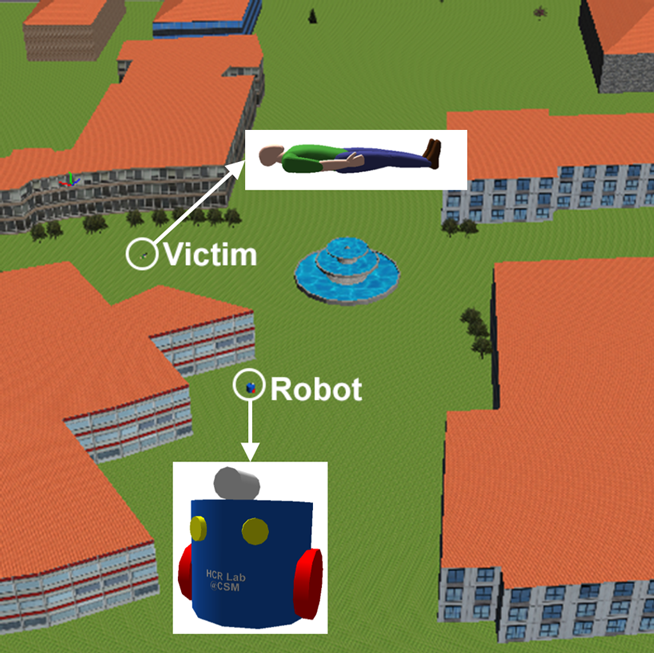}
    \end{minipage}}
  \subfigure[Google map of real scenes]{
     \label{fig:csm_google_map}%% label for first subfigure
    \begin{minipage}[b]{0.19\textwidth}
      \centering
        \includegraphics[height = 1.475in]{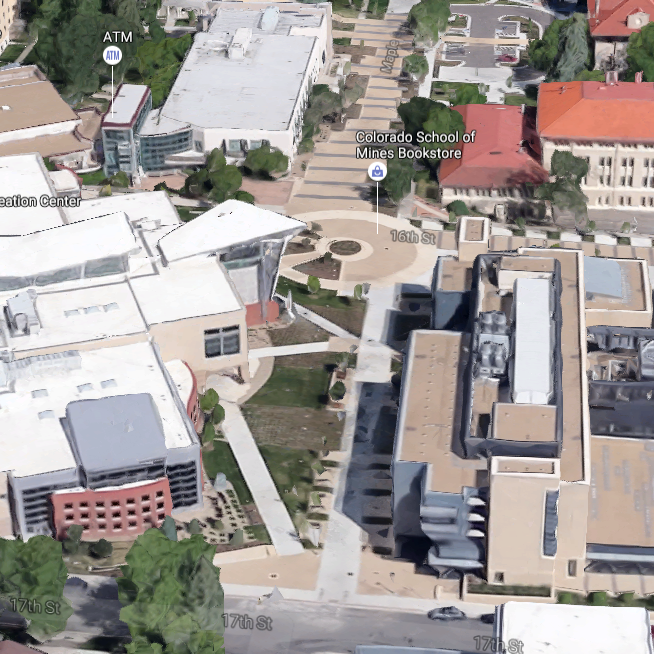}
    \end{minipage}}
  \subfigure[Rescuebot moving route and acquired observations]{
    \label{fig:webots_route} %% label for second subfigure
    \begin{minipage}[b]{0.6\textwidth}
      \centering
        \includegraphics[height = 1.475in]{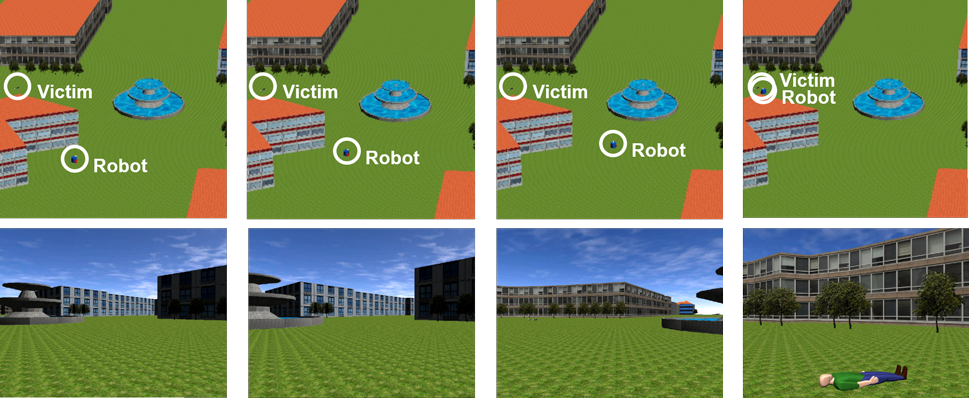}
    \end{minipage}}
  \caption{
  Experiment setups and qualitative results in robot-assisted urban search and rescue scenarios.
  Fig. \ref{fig:csm_webots_map} illustrates the simulated environment.
   Fig. \ref{fig:csm_webots_map} shows the Google satellite map of the real campus environment of the Colorado School of Mines.  Fig. \ref{fig:webots_route} illustrates qualitative results with the top row showing the robot moving route and the bottom row showing the observations obtained by the robot camera.
}
\label{fig:csm_maps} %% label for entire figure
\end{figure*}

\subsection{Urban Search and Rescue Simulation}
In this set of experiments, we apply the Webots simulator\footnote{Webots: \url{https://www.cyberbotics.com}.}  \cite{michel2004webotstm} to evaluate our SMAL approach in an urban search and rescue application.
The objective is to let a robot learn how to find victims in large urban areas from expert demonstrations.
We chose the campus of the Colorado School of Mines as our urban environments. The Google satellite map of this area is shown in Fig. \ref{fig:csm_google_map}.
We imported the OpenStreetMap\footnote{OpenStreetMap: \url{http://www.openstreetmap.org/#map=18/39.74966/-105.22212}.}
of this area into the Webots platform,
as illustrated in Fig. \ref{fig:csm_webots_map}.
The robot and victim models we built in the Webots platform
are shown in Fig. \ref{fig:csm_webots_map}.
The two-wheel mobile robot, named \emph{Rescuebot}, equips with a color camera with a $1024\times 768$ resolution.
In addition, we are able to obtain the accurate Rescuebot's location and rotation information from the simulator, which is used as the ground truth to evaluate state recognition.
The victim is lying on the ground without any movement during the entire simulation period, waiting for a robot to find him.

During the training process, we teleoperated the \emph{Rescuebot} to approach the target victim using keyboards as the expert demonstration.
The image sequences obtained by the \emph{Rescuebot} and the keyboard teleoperation commands were recorded to train our SMAL method.
After training was completed, the \emph{Rescuebot} was able to automatically execute search operations using the learned model
in the testing phase.

To qualitatively evaluate the experimental results, an example route that the \emph{Rescuebot} successfully finds the victim in the execution phase is presented in Fig. \ref{fig:webots_route}.
It demonstrates that, although the \emph{Rescuebot} cannot see the victim directly,
the robot is still able to move and search around to locate the victim.
This qualitative result demonstrates that our SMAL method
enables robots to learn how to autonomously search for victims in urban search and rescue scenarios.

\begin{figure}[htp]
\vspace{-3pt}
  \subfigure[Precision-recall curves]{
     \label{fig:pr_webots}%% label for first subfigure
    \begin{minipage}[b]{0.23\textwidth}
      \centering
        \includegraphics[width=1\textwidth]{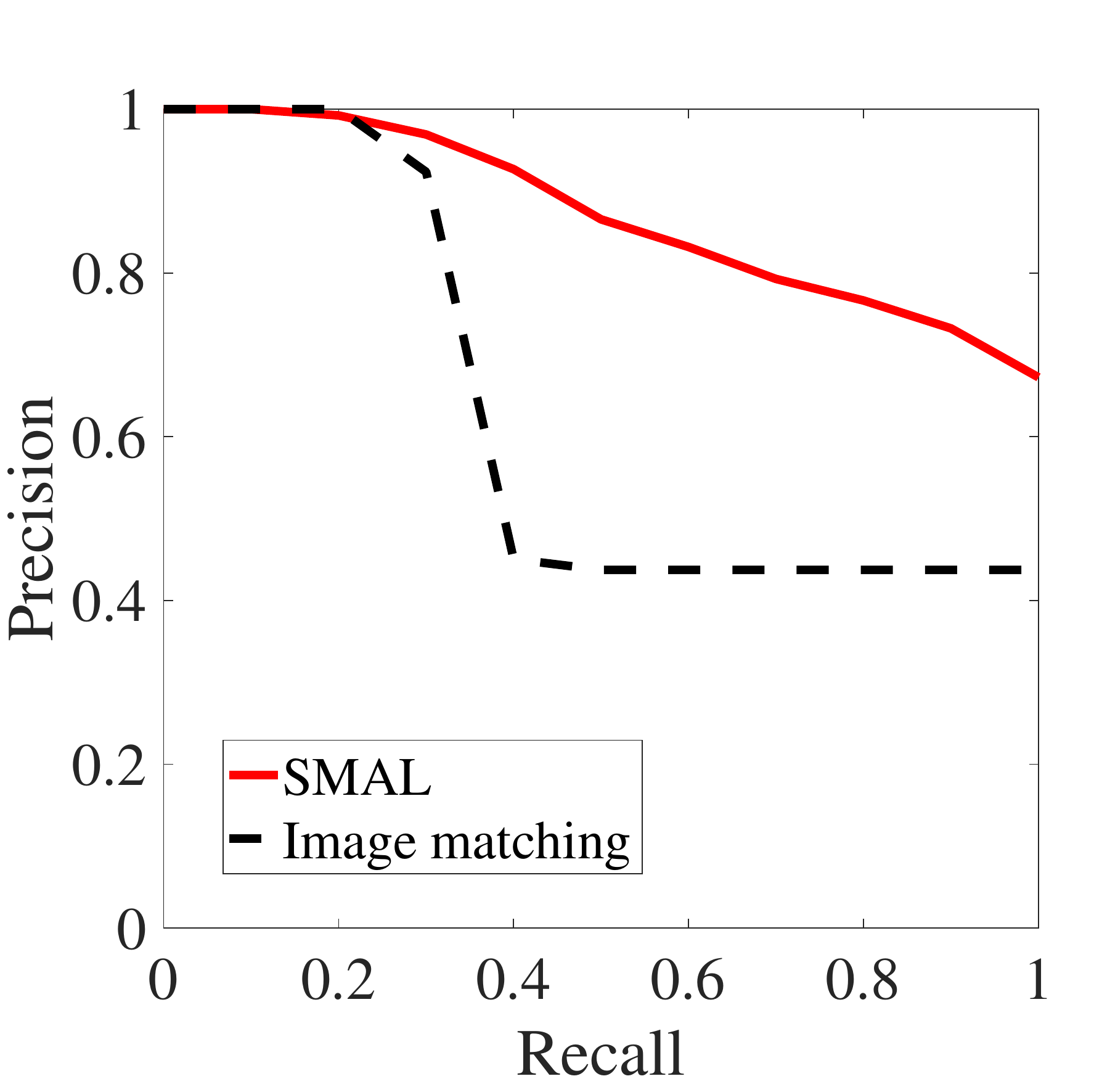}
    \end{minipage}}
%  \hspace{1pt}
  \subfigure[Objective value]{
    \label{fig:obj_webots} %% label for second subfigure
    \begin{minipage}[b]{0.23\textwidth}
      \centering
        \includegraphics[width=1\textwidth]{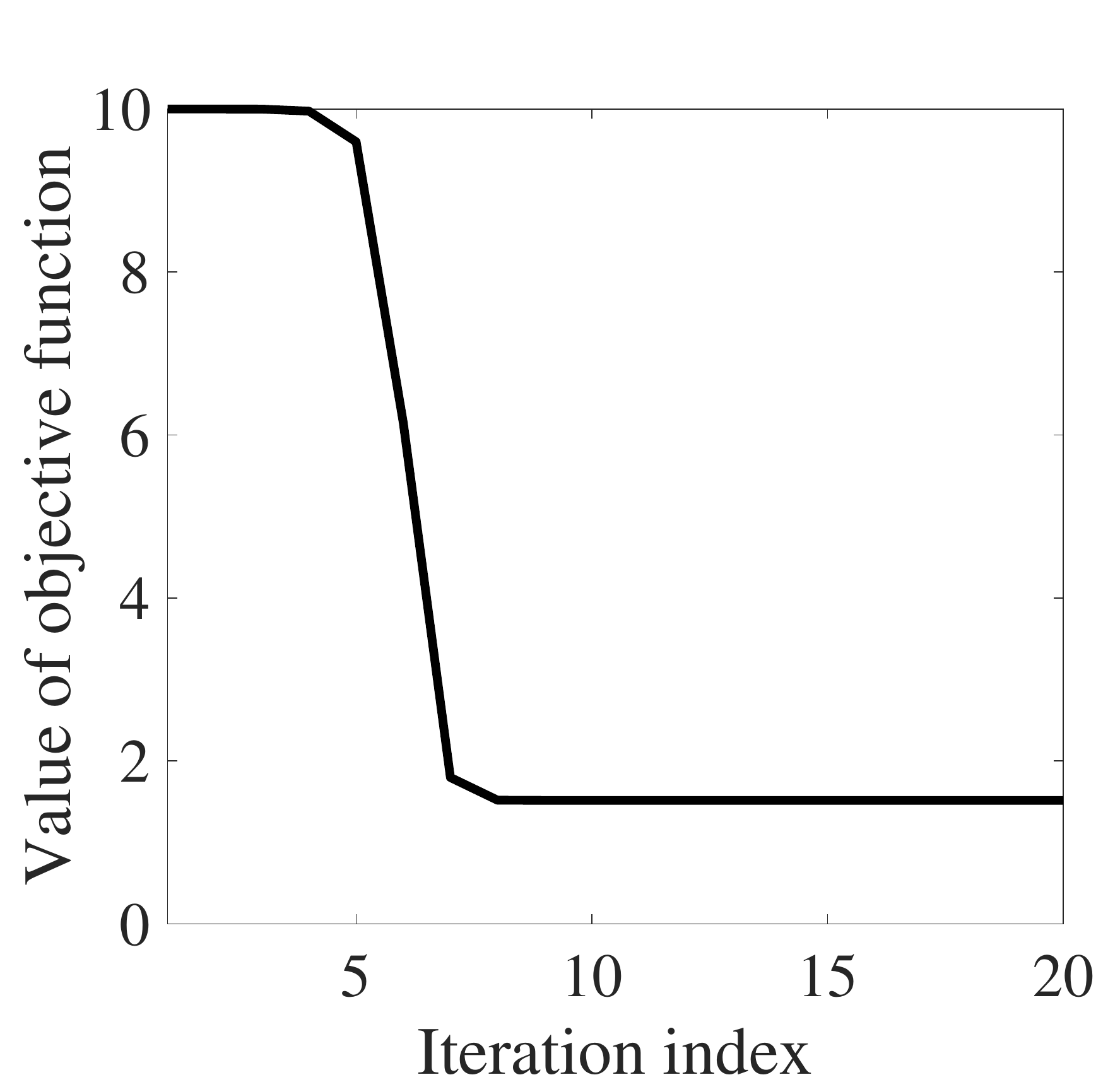}
    \end{minipage}}
  \caption{
Quantitative evaluation of our SMAL approach in simulated urban search and rescue scenarios.
}
\label{fig:quan_webots} %% label for entire figure
\end{figure}

In addition, we perform quantitative validation using the precision-recall curve as a metric to evaluate the performance of state recognition,
as shown in Fig. \ref{fig:pr_webots} (curves closer to the top right corner indicating a better performance).
We also compared the SMAL approach to the baseline method based on individual images with the same modalities,
which is demonstrated in Fig. \ref{fig:pr_webots}.
It is observed that our SMAL method for sequence-based state recognition
outperforms the baseline method using individual images.

We also evaluate the efficiency of our methods for state recognition through studying the value of objective function iteratively updated by
Algorithm \ref{alg:OptAlgorithm}.
The result, presented in Fig. \ref{fig:obj_webots},
indicates the algorithm converges in 9 iterations (in general, it converges within 20 iterations with the value below $10^{-4}$,
which demonstrates the algorithm efficiency to solve the formulated regularized optimization problem.

\begin{figure*}[htb]
  \subfigure[Experiment setup]{
     \label{fig:lab_map}%% label for first subfigure
    \begin{minipage}[b]{0.28\textwidth}
      \centering
        \includegraphics[height=1.5in]{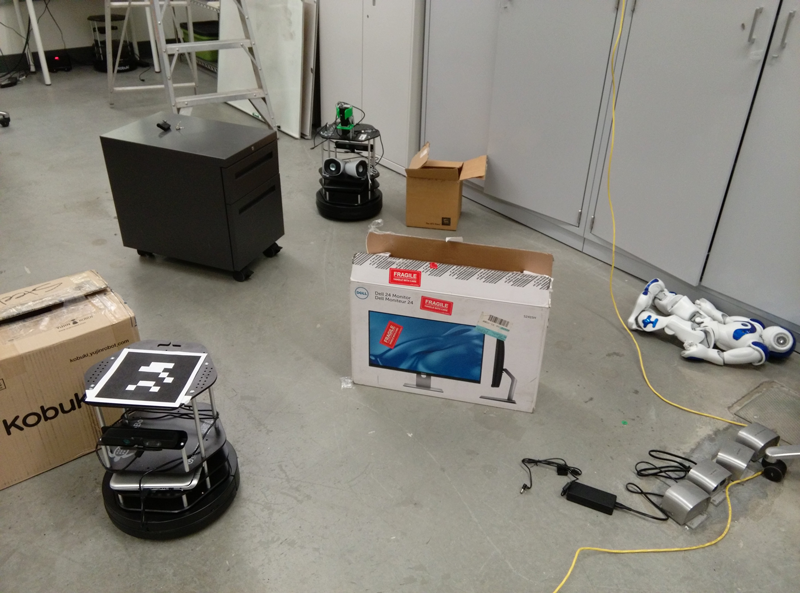}
    \end{minipage}}
  \hspace{1pt}
  \subfigure[Moving route and observations of the TurtleBot during execution.]{
    \label{fig:lab_route} %% label for second subfigure
    \begin{minipage}[b]{0.65\textwidth}
      \centering
        \includegraphics[height=1.5in]{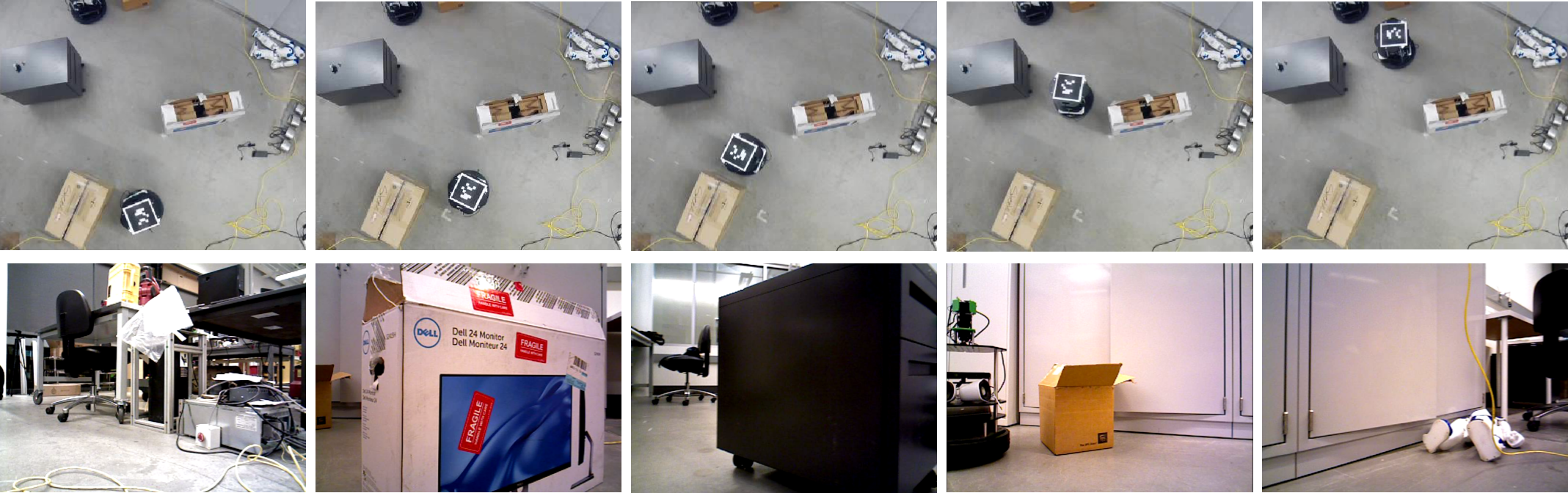}
    \end{minipage}}
  \caption{
Experiment setups of robot-assisted search and rescue in indoor environments and qualitative results.
Fig. \ref{fig:lab_map} shows the indoor environment used in this set of experiments for robots to search the victim (i.e., the NAO robot).
Qualitative experimental results are presented in Fig. \ref{fig:lab_route}, with the top row showing the moving route of the TurtleBot from the viewpoint of an overhead camera, and the bottom row showing the observations acquired by the TurtleBot during the execution.
}
\label{fig:lab_maps} %% label for entire figure
\end{figure*}

\subsection{Indoor Search and Rescue using Real TurtleBot}

In this set of experiments,
we evaluate our SMAL method to teach robots to perform victim search in indoor scenarios.
A real TurtleBot II robot is used to evaluate the performance of our system.
The objective is to teach the TurtleBot about how to find victims (in this experiment, a NAO humanoid robot) in the room using expert demonstrations.
The setup of the indoor search area is presented in Fig. \ref{fig:lab_map}.
We also install an overhead camera above this area to collect the ground truth of robot location and orientation for evaluation only by tracking the ARTag attached on top of the Turtlebot.

In the training phase, we teleoperated the TurtleBot using keyboards as demonstrations to let it approach the Nao robot.
The observation obtained by the TurtleBot and the keyboard teleoperation commands were recorded to train our SMAL model.
After that, during the execution phase, the TurtleBot executed the search task based on the learned model to find the NAO robot.
A challenge of this real-world experiment in comparison to simulation is that the TurtleBot often shook when moving, making the captured observations unstable, which can decrease the accuracy of state recognition.

\begin{figure}[bp]
\vspace{-4pt}
  \subfigure[Precision-recall curve]{
     \label{fig:pr_lab}%% label for first subfigure
    \begin{minipage}[b]{0.23\textwidth}
      \centering
        \includegraphics[width=1\textwidth]{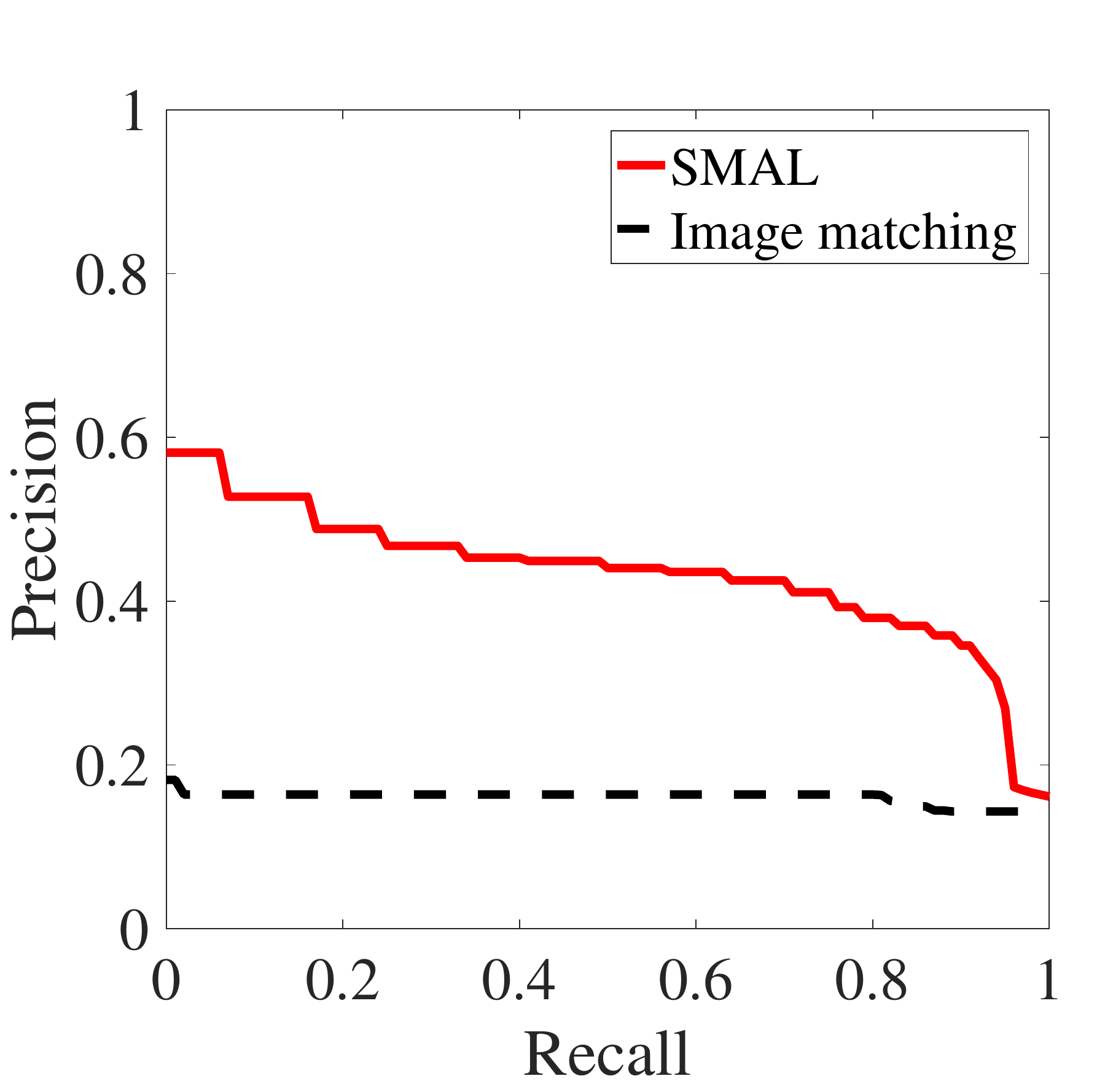}
    \end{minipage}}
%  \hspace{1pt}
  \subfigure[Objective value]{
    \label{fig:obj_lab} %% label for second subfigure
    \begin{minipage}[b]{0.23\textwidth}
      \centering
        \includegraphics[width=1\textwidth]{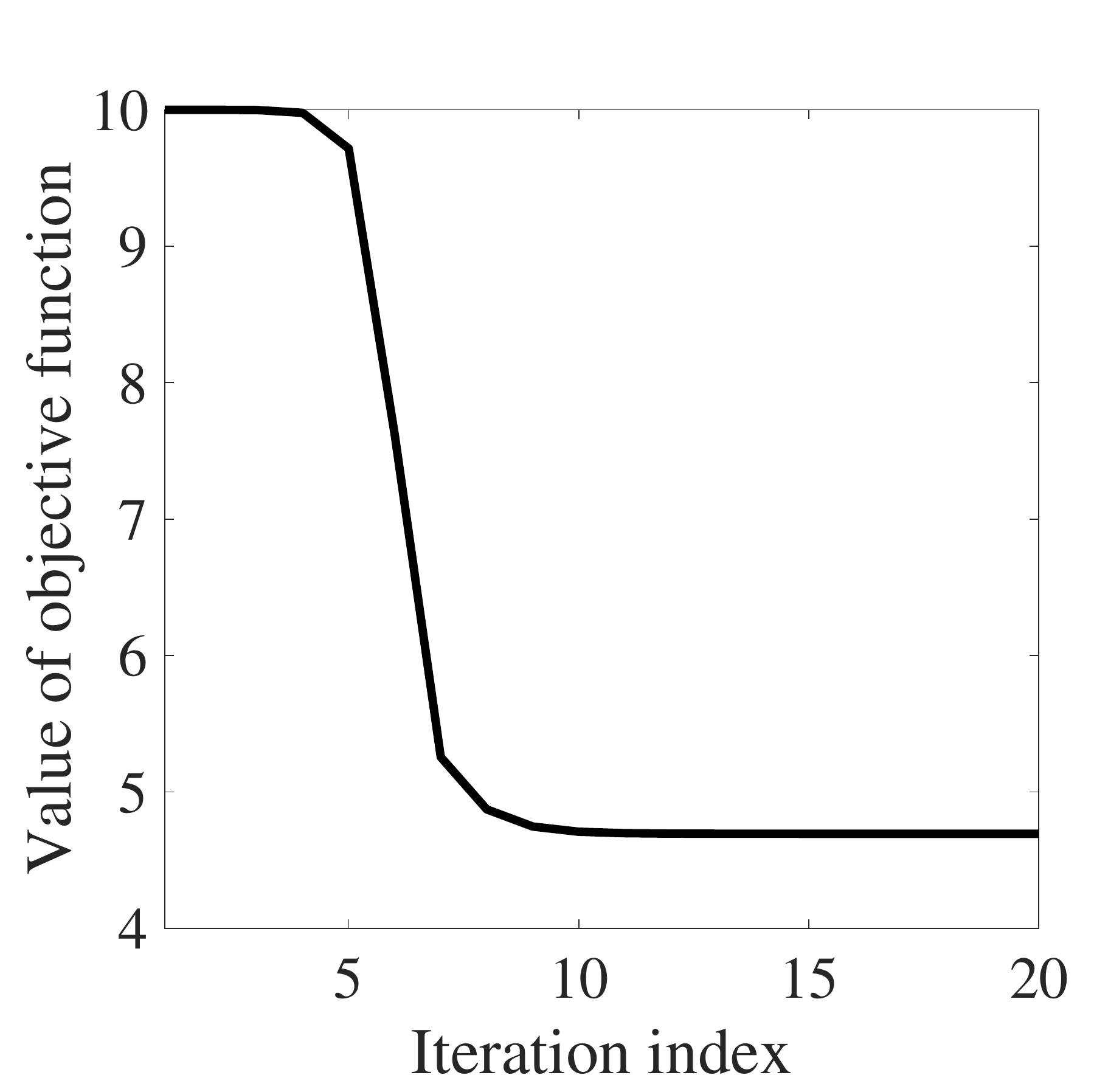}
    \end{minipage}}
  \caption{
Quantitative evaluation of our SMAL approach in real-world indoor search and rescue scenarios.}
\label{fig:quan_lab} %% label for entire figure
\end{figure}

The qualitative experimental results are illustrated in Fig. \ref{fig:lab_route},
which indicates even the Turtlebot cannot directly see the victim (i.e., the NAO robot in this set of experiments),
but it can still navigate around multiple obstacles to find the victim.
This demonstrates the effectiveness of our SMAL approach to teach robots about how to search victims in a real indoor environment.
We also quantitatively evaluate our method's performance using precession-recall curves and compare SMAL with the baseline method using image matching.
The results are presented in Fig. \ref{fig:pr_lab},
which shows our approach significantly outperforms the baseline method.
The efficiency of our SMAL approach is proved in Fig. \ref{fig:obj_lab},
which shows the algorithm converges after 12 iterations.

\begin{figure}[htp]
\vspace{-5pt}
  \subfigure[Precision-recall curves]{
     \label{fig:pr_lab_all}%% label for first subfigure
    \begin{minipage}[b]{0.23\textwidth}
      \centering
        \includegraphics[width=1\textwidth]{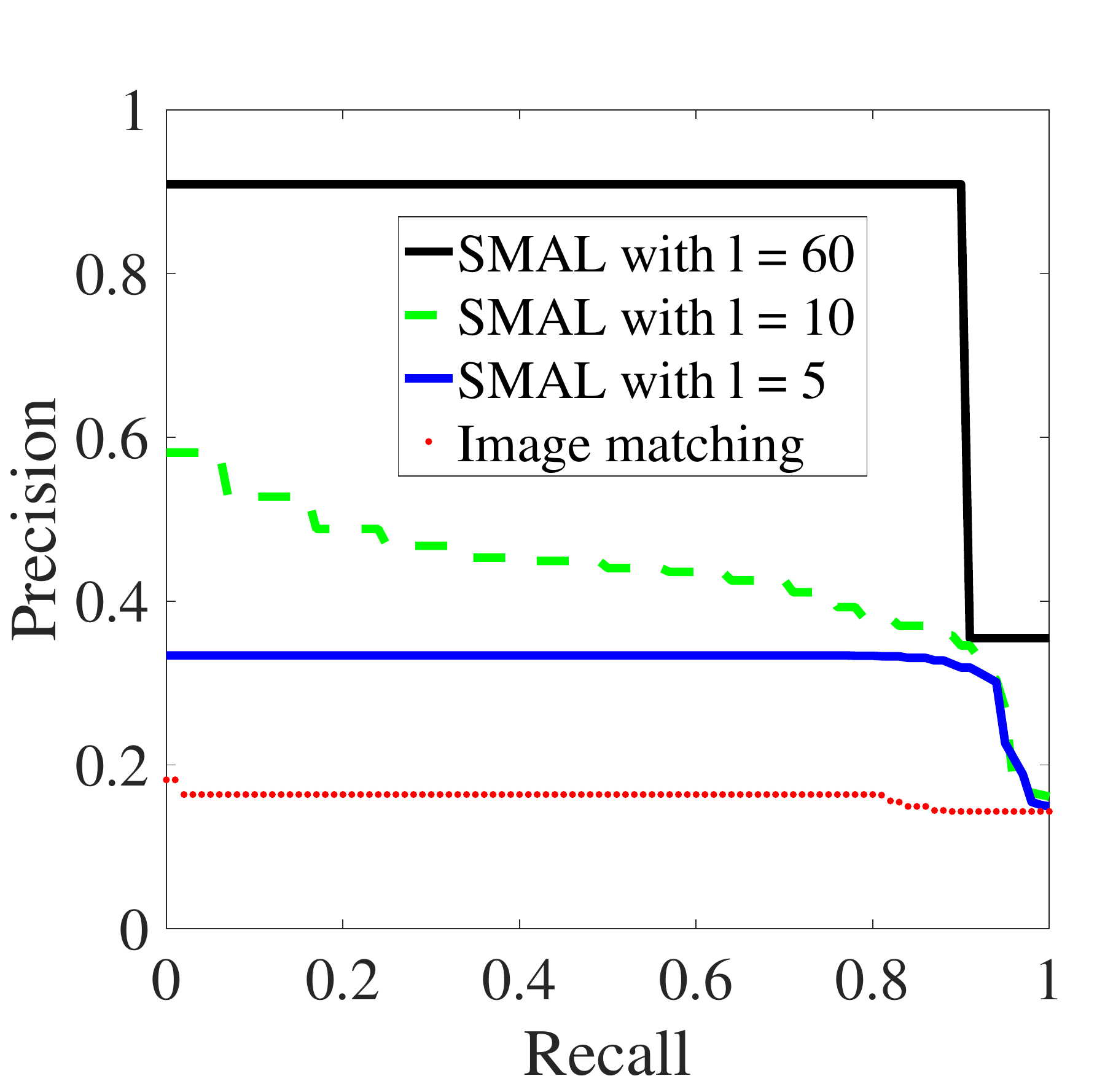}
    \end{minipage}}
%  \hspace{1pt}
  \subfigure[Success rates]{
    \label{fig:bar} %% label for second subfigure
    \begin{minipage}[b]{0.23\textwidth}
      \centering
        \includegraphics[width=0.92\textwidth]{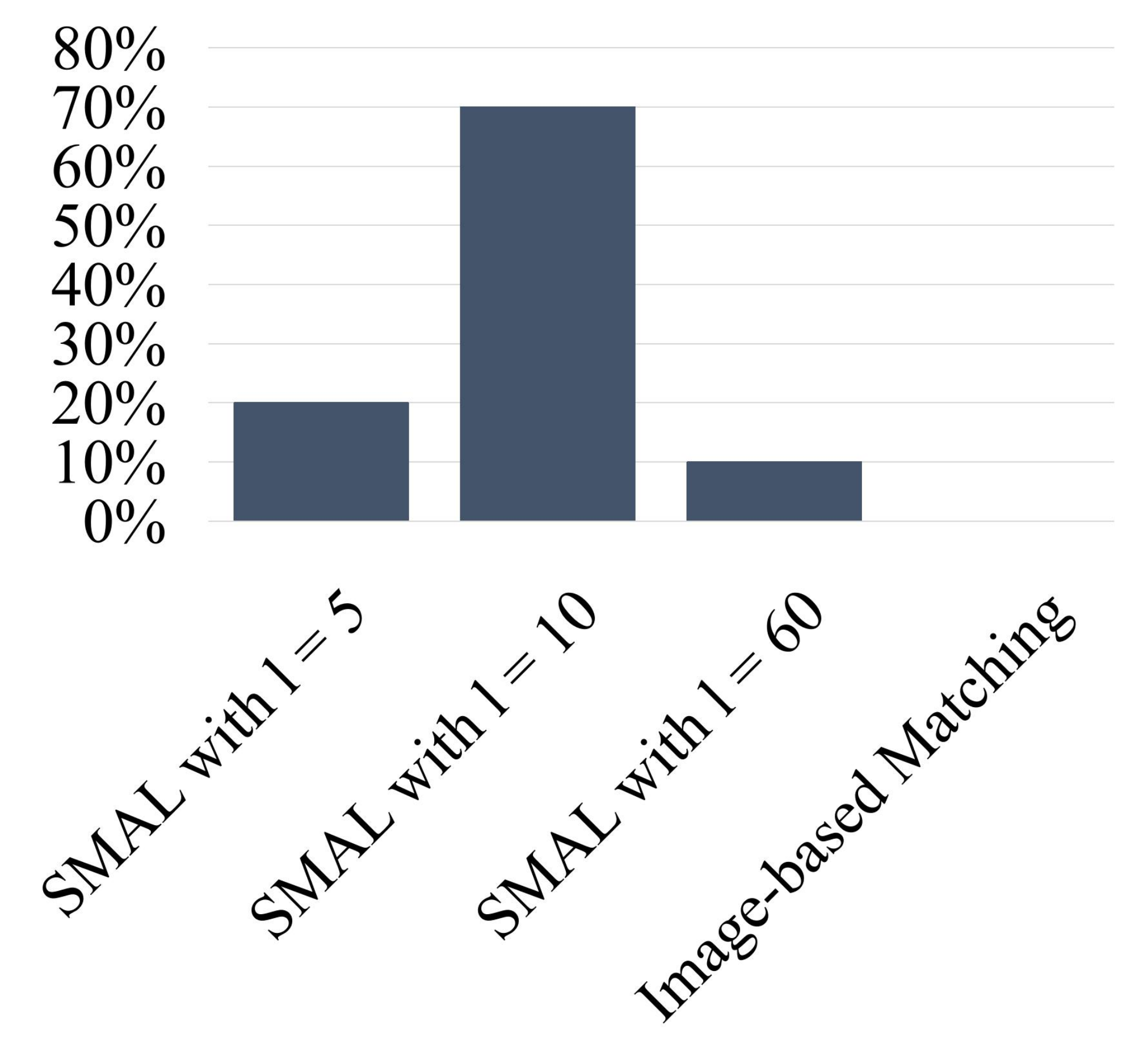}
    \end{minipage}}
  \caption{
  Performance evaluation of SMAL using different parameter values.
}
\label{fig:analysis} %% label for entire figure
\end{figure}

\subsection{Parameter Analysis}

We analyze the effects of various parameter values on our SMAL approach in real-world indoor search tasks using real TurtleBots.

The sequence length $l$ for state recognition is the most important parameter.
The precision-recall curves in Fig. \ref{fig:pr_lab_all} indicate that
better performance can be obtained when we increase the sequence length.
That is because long sequences can provide more comprehensive information than short sequences.
When $l=1$, a sequence becomes a single image.
In addition, we use success rate as a metric to evaluate the percentage that the robot can successfully find victims without hitting obstacles.
The results are demonstrated in Fig. \ref{fig:bar}, where 10 executions are used in each case to calculate the success rate.
It is observed that when the used sequence is short,
the poor perception result negatively affects decision making,
resulting in the low success rate.
However, longer sequences do not necessarily result in higher success rates.
%Fig. \ref{fig:bar} shows that when $s=60$ is much lower than that when $s=10$.
This is because as we increase the sequence length,
although each sequence can contain more information,
the frequency in which the robot receives observations decreases.
This can dramatically decrease the succuss rate,
since the information does not come in time for robot control.